\documentclass[letterpaper]{article} 
\usepackage{aaai24}  
\usepackage{times}  
\usepackage{helvet}  
\usepackage{courier}  
\usepackage[hyphens]{url}  
\usepackage{graphicx} 
\usepackage{natbib}  
\usepackage{caption} 
\usepackage{algorithm}
\usepackage{algorithmic}
\usepackage{amsmath}
\usepackage{newfloat}
\usepackage{listings}
\usepackage{booktabs}
\usepackage{multirow}
\newcommand{\rotbox}[1]{\rotatebox{40}{#1}}

\usepackage{subfigure}
\usepackage{amsfonts}
\usepackage[misc]{ifsym}

\newcommand\blfootnote[1]{%
\begingroup
\renewcommand\thefootnote{}\footnote{#1}%
\addtocounter{footnote}{-1}%
\endgroup
}
\DeclareCaptionStyle{ruled}{labelfont=normalfont,labelsep=colon,strut=off} 
\floatstyle{ruled}
\newfloat{listing}{tb}{lst}{}
\floatname{listing}{Listing}
\title{DPL: Decoupled Prompt Learning for Vision-Language Models}

\author {
    Chen Xu\equalcontrib\textsuperscript{\rm 1},
    Yuhan Zhu\equalcontrib\textsuperscript{\rm 1}\thanks{Work is done during internship at  vivo AI Lab.},
    Guozhen Zhang\textsuperscript{\rm 1},
    Haocheng Shen\textsuperscript{\rm 2},
    Yixuan Liao\textsuperscript{\rm 2},
    Xiaoxin Chen\textsuperscript{\rm 2},
    Gangshan Wu\textsuperscript{\rm 1},
    Limin Wang\textsuperscript{\rm 1,~\Letter}
}
\affiliations {
    \textsuperscript{\rm 1}State Key Laboratory for Novel Software Technology, Nanjing University\\
    \textsuperscript{\rm 2}vivo AI Lab\\
}

\usepackage{bibentry}

\begin{document}

\maketitle

\begin{abstract}
Prompt learning has emerged as an efficient and effective approach for transferring foundational Vision-Language Models (e.g., CLIP) to downstream tasks. However, current methods tend to overfit to seen categories, thereby limiting their generalization ability for unseen classes. In this paper, we propose a new method, Decoupled Prompt Learning (DPL), which reformulates the attention in prompt learning to alleviate this problem. Specifically, we theoretically investigate the collaborative process between prompts and instances (i.e., image patches/text tokens) by reformulating the original self-attention into four separate sub-processes. Through detailed analysis, we observe that certain sub-processes can be strengthened to bolster robustness and generalizability by some approximation techniques. Furthermore, we introduce language-conditioned textual prompting based on decoupled attention to naturally preserve the generalization of text input. Our approach is flexible for both visual and textual modalities, making it easily extendable to multi-modal prompt learning. By combining the proposed techniques, our approach achieves state-of-the-art performance on three representative benchmarks encompassing 15 image recognition datasets, while maintaining parameter-efficient. Moreover, our DPL does not rely on any auxiliary regularization task or extra training data, further demonstrating its remarkable generalization ability.
\end{abstract}

\blfootnote{ \Letter: Corresponding author (lmwang@nju.edu.cn).}

\section{Introduction}

Learning from the paired text and image data, Vision-Language Models (VLMs) such as CLIP~\cite{CLIP} have demonstrated remarkable generalization capabilities and robustness across different domains and downstream tasks. These methods learn to align visual and textual representations in the same feature space using separate visual and textual encoders. Pre-trained on a vast dataset of image-text pairs, CLIP exhibits high-quality zero-shot recognition ability on downstream tasks. During inference, class names are combined with hand-crafted templates (e.g., ``\texttt{a photo of a}'') and fed into the textual encoder to produce class-specific embeddings. Classification scores are then computed as cosine similarities between these textual embeddings and the visual embeddings from visual encoders. 


\begin{figure}[t]
\centering\includegraphics[width=0.48\textwidth]{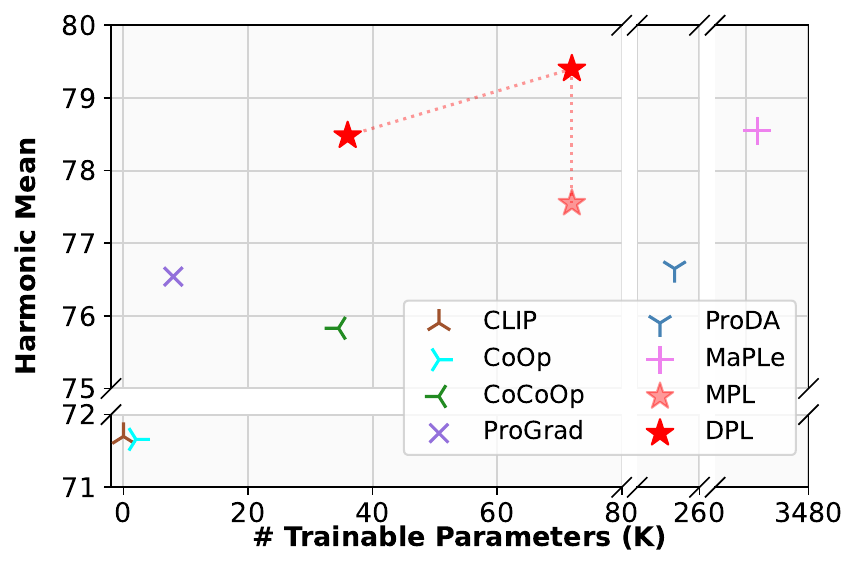}
    \caption{ \textbf{Comparison with previous prompt learning methods on the average harmonic mean in the base-to-new setting}. DPL surpasses all previous methods by a significant margin while requiring considerably fewer parameters than the previous SOTA. MPL refers to vanilla multi-modal prompt learning.}
    \label{fig:params-hm}
\end{figure}

Recently, inspired by the success in NLP, prompt learning has emerged as an effective approach to adapt VLMs to downstream tasks~\cite{CoOp,CoCoOp,ProDA}. By replacing hand-crafted templates with learnable embeddings and keeping the model weights frozen, prompt learning enables the model to efficiently learn from downstream tasks and achieve excellent performance in few-shot scenarios~\cite{CoOp}. Despite these advantages, such methods often confront the common issue of overfitting to seen classes~\cite{CoCoOp}. As a consequence, performance on unseen classes drops significantly compared to the original pre-trained model. Recent studies have attempted to alleviate this issue by introducing additional auxiliary tasks such as aligning learned prompts closely with hand-crafted templates by using cross-entropy loss~\cite{LASP} or employing extra learnable modules, e.g., M-Net in CoCoOp~\cite{CoCoOp} and cross-modal projection layers in MaPLe~\cite{MaPLe}. 


Unlike previous approaches, in this paper, we focus on decoupling the computation structure in prompt learning and propose a novel method to relieve the overfitting issue. First, we theoretically reformulate and divide the attention process in vanilla prompt learning into four decoupled sub-processes from two perspectives: instance forwarding and prompt forwarding. We then provide detailed explanations for their own function and identify that certain sub-processes can be reinforced to enhance the generalization ability and robustness of the learned prompts. Based on these findings, we propose a novel method called \textbf{D}ecoupled \textbf{P}rompt \textbf{L}earning (\textbf{DPL}) which recombines the sub-processes in a decoupled manner, resulting in a more generalizable and robust framework for prompts to learn from downstream tasks. We empirically remove the interaction between prompt tokens and provide intuitive speculations and interpretations for this approximation. To further enhance the generalization ability of the text input, we introduce language-conditioned textual prompting, which leverages the prior knowledge from hand-crafted templates and preserves the knowledge well with our proposed decoupled attention process. Through comprehensive experiments on three crucial representative settings, including base-to-new generalization, cross-dataset generalization, and domain generalization, DPL surpasses previous state-of-the-art methods on all three benchmarks with fewer parameters (e.g., Fig.~\ref{fig:params-hm}). Moreover, our DPL does not rely on any additional trainable modules, auxiliary regularization tasks, or extra training data, further demonstrating its superior generalizability and robustness. The main contributions of our work are summarized as follows:
\begin{enumerate}
    \item We are the first to explore how to improve the computational structure of prompt learning in VLMs. We present an intriguing view on the attention computation of vanilla prompt learning by decomposing it into four sub-processes and identifying the superfluous coupled components which may have an adverse impact on the generalization capability of the pre-trained model.
    \item To address these issues, we propose a novel method called Decoupled Prompt Learning (DPL), which effectively decouples the attention process in prompt learning and minimizes unnecessary information interaction that can lead to sub-optimal performance.
    \item DPL achieves the state-of-the-art performance on three key representative benchmarks encompassing 15 image recognition datasets, while maintaining parameter-efficient. This verifies the remarkable generalizability of DPL without auxiliary regularization tasks, extra modules, or additional data.
\end{enumerate}

\section{Related Work}
\noindent\textbf{Vision-Language Models.} 
Large Vision-Language Models (VLMs) leverage the power of learning joint representations by combine visual and linguistic data during the pre-training stage. These methods have demonstrated remarkable capabilities across a diverse set of challenging tasks. Leading works, such as CLIP~\cite{CLIP}, ALIGN~\cite{jia2021scaling}, Florence~\cite{yuan2021florence}, LiT~\cite{zhai2022lit}, DeCLIP~\cite{li2021supervision}, bridge the gap between vision and language modalities by employing contrastive learning strategies. These techniques enable the model to acquire image and text representations in the same feature space, leveraging the learning on large corpora of text-image pairs (e.g., $\sim$400M for CLIP and $\sim$ 1B for ALIGN). In this study, we build upon CLIP as the foundation model of our approach and focus on its adaption techniques.
 
\noindent\textbf{Prompt Learning.} In Natural Language Processing (NLP) tasks, the introduction of instructive sentences tailored to specific tasks as extra input has been shown to enhance the language model's ability to generate more contextually relevant outputs~\cite{pet}. These sentences, known as text prompts, were initially handcrafted for each individual task. Later works \cite{PromptTuning,li2021prefix,P-tuning} replaced handcrafted words with continuous token embeddings and optimized these learnable prompts during training, which is known as ``Prompt Learning''. This methodology has demonstrated promising performance in related fields. The technique was extended to the visual modality by Visual Prompt Tuning \cite{VPT}, which inserts learnable prompts at each layer of ViT\cite{VPT} to facilitate the adaptation process and confirm its effectiveness. 

\noindent\textbf{Prompt Learning in Vision Language Models.} When applying VLMs (e.g., CLIP~\cite{CLIP}) to downstream tasks, it is often necessary to design appropriate textual prompts. However, this process can be labor-intensive and sub-optimal. To address this issue, CoOp~\cite{CoOp} proposed to learn text prompts in continuous space and achieve promising results. However, CoCoOp~\cite{CoCoOp} showed that the learned prompts in the downstream domain are severely overfitted to the seen categories, resulting in poor performance on unseen categories. This undermines the most valuable advantage of CLIP, which is the ability for open-vocabulary classification. To alleviate this problem, follow-up efforts have introduced additional network modules that require more trainable parameters~\cite{CoCoOp,ProDA,MaPLe}, or auxiliary tasks~\cite{ProGrad,KgCoOp,PLOT} such as regularizing the learning process using the original CLIP. Other approaches have used additional data~\cite{MVLPT,POMP}, or require additional training at test-time~\cite{TPT,RLCF}. Our proposed DPL investigates the inherent structure of prompt learning and proposes to improve the collaborative process between prompts and instances.



\begin{figure*}[t]
    \centering
\includegraphics[width=0.98\textwidth]{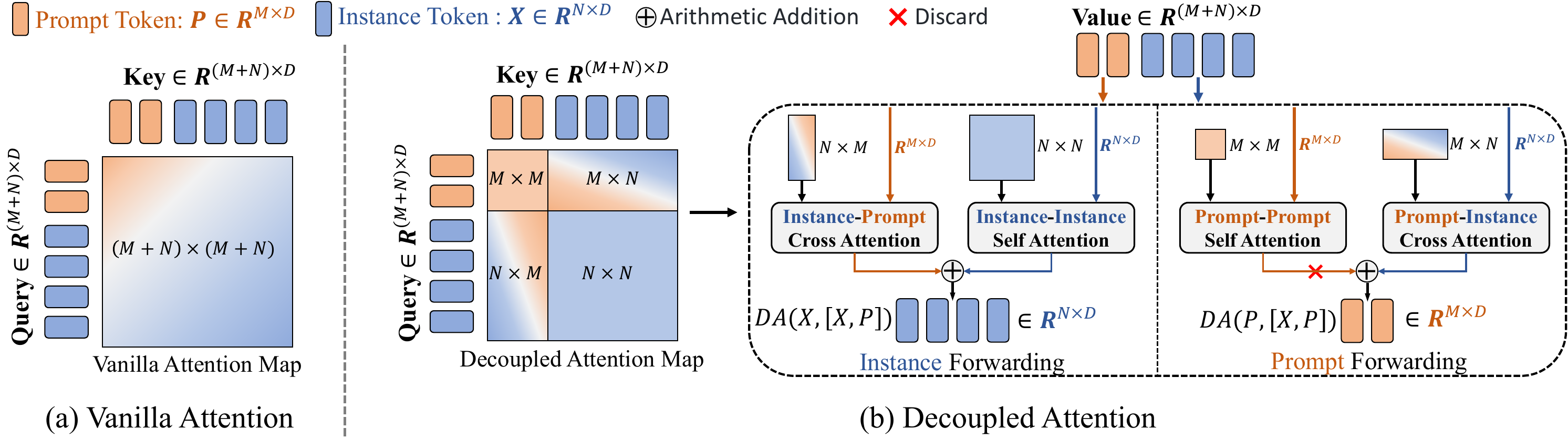}
    \caption{\textbf{(a) Attention map in vanilla prompt learning. (b) Our decoupled attention for prompt learning.} We separate the self-attention within $[X,P]$ into four decoupled sub-processes, and discard the uninterpretable interaction between prompts.}
    \label{fig:main_arch}
\end{figure*}

\section{Method}

\subsection{Prompt Learning Revisited}
Prompt learning was first introduced in \cite{VPT} for computer vision tasks, aiming to efficiently and effectively transfer pre-trained foundational models to downstream tasks. Given a pre-trained Transformer model, learnable prompt tokens are inserted to $i$-th layer ($L_i$) while the entire model is kept frozen. We denote the original input instance embeddings and the newly introduced learnable prompts by $X \in \mathbb{R}^{N\times D}$ and $P\in \mathbb{R}^{M\times D}$, respectively. The forward pass of this layer can be formulated as:
\begin{equation}
\begin{aligned}
{\left[O_{i+1}, X_{i+1}\right] } & =L_i\left(\left[P_{i}, X_{i}\right]\right),
\label{equ:vpt_forward}
\end{aligned}
\end{equation}
where $\left[\cdot, \cdot\right]$ represents the stacking operation along the length of the sequence. $O_{i+1}$ is the output of corresponding prompt tokens, which can either be discarded~\cite{VPT,MaPLe}, or further utilized by subsequent layers~\cite{CoOp, ProVP}. For the sake of simplicity, we omit the layer index below.

\subsection{Reformulation of Attention in Prompt Learning}
For a general attention process, for any query embedding $y_i \in Y$ as query and $Z$ as key set, we denote the denominator in the \texttt{softmax} function during the calculation of the attention scores as:
\begin{equation}
\begin{aligned}
\lambda_i(Y,Z)=\lambda(y_i,Z)=\sum\limits_{z\in Z}e^{q(y_i)k(z)},
\end{aligned}
\end{equation}
the corresponding attention output of $y_i$ is denoted by $\mathcal{A}_i(Y,Z)=\mathcal{A}(y_i,Z) = e^{q(y_i)k(Z)}\times v(Z)/\lambda_i(Y,Z)$, where $q, k, v$ are the query, key and value projection functions.

When learnable prompts $P$ are incorporated, the self-attention process within $[X,P]$ can be viewed as a self-attention within $X$, where the denominator in the \texttt{softmax} function is influenced by the introduced prompts and an additional cross-attention to the prompts:
\begin{equation}
\begin{aligned}
\mathcal{A}_i(X,[X,P])=f_i(X, P)\mathcal{A}_i(X,X)+h_i(X, P)\mathcal{A}_i(X,P),
\label{equ.OrignalFunc_i}
\end{aligned}
\end{equation}
where $f_i(X,P)+h_i(X,P)=1$ always holds, and we define $f_i$ and $h_i$ as the multiplier of variation of the influenced denominator:
\begin{equation}
\begin{aligned}
f_i(X,P)&=\frac{\lambda_i(X,X)}{\lambda_i(X,[X,P])}\\
h_i(X,P)&=\frac{\lambda_i(X,P)}{\lambda_i(X,[X,P])}.\\
\end{aligned}
\end{equation}

Then we omit the subscript $i$ to represent the \textit{instance forwarding process} of $X$ in vanilla prompt learning
\begin{equation}
\begin{aligned}
\mathcal{A}(X,[X,P])=f(X, P)\mathcal{A}(X,X)+h(X, P)\mathcal{A}(X,P).
\end{aligned}
\label{equ.OrignalFunc}
\end{equation}
Similarly, for prompt tokens $P$, the \textit{prompt forward process} can be expressed as
\begin{equation}
\mathcal{A}(P,[X,P])=f(P,X)\mathcal{A}(P,P)+h(P,X)\mathcal{A}(P,X).
\label{equ.PFunc}
\end{equation}

In Eq.~\ref{equ.OrignalFunc} and Eq.~\ref{equ.PFunc}, the original attention result $\mathcal{A}([X,P],[X,P])$ in vanilla prompt learning can be separated into four sub-processes that are coupled with each other by their coefficients $f$ and $h$.

\subsection{Decoupled Prompt Learning}
In this section, we introduce a novel approach called Decoupled Prompt Learning (DPL), aiming at improving the robustness of prompts. Through a detailed analysis of the effectiveness of the four attention sub-processes and we reveal that each of them serves distinct objectives and that certain terms can be reinforced to bolster the robustness and generalization ability without compromising adaption capability. Based on these insights, we introduce DPL, which effectively decouples the overall attention by recombining the decoupled sub-processes to form a more effective and generalizable optimization process in the downstream tasks.


\noindent \textbf{Instance Forwarding.} We consider two extreme situations from Eq.~\ref{equ.OrignalFunc}: if $f_i(X,P)=1$, the instance forwarding degrades to zero-shot CLIP, which blocks learning from downstream tasks as $h_i(X,P)=0$. On the other hand, if $h_i(X, P) = 1$, the output relies solely on the information extracted from the learned prompts, disregarding the original self-attention result of pre-trained CLIP. Reflecting on these observations, we intuitively assume that the second term is responsible for achieving adaptation goals, whereas the first term, which shares a computational process akin to zero-shot CLIP, could contribute to preserving the generalization ability acquired during the pre-training stage.

Inspired by the above intuition, to emphasize the generalization ability of the model, we focus on analyzing the case where $f_i(X, P)$ close to 1 for 
$\forall x_i \in X$. Under this assumption, the following limit always holds:
\begin{equation}
    \frac{h_i(X,P)}{f_i(X,P)}=\frac{\lambda_i(X,P)}{\lambda_i(X,X)}=\frac{\sum\limits_{p_j\in P}e^{q(x_i)k(p_j)}}{\sum\limits_{x_j\in X}e^{q(x_i)k(x_j)}}\rightarrow 0.
\end{equation}
This indicates that, after the projections, for a fixed $N$ and $M$, the learned prompts exhibit relatively low similarities with the instance embeddings from the training set. This suggests that such prompts are prevented from forming similar representations to the output distribution of the training subset, allowing them to be less influenced by potential shifts between the testing and training samples in the feature representation space. As a result, they could avoid excessive and over adaptation to downstream datasets and focus more on capturing task-specific information from a global perspective. This characteristic signifies that such learned prompts are more robust and generalizable. 


To leverage the aforementioned advantages, we incorporate the assumption that $f_i(X,P)=1$ into the instance forwarding process in an approximate manner, while $h_i(X,P)$ is fairly replaced with an approximate value to the scaled result, denoted as $\sigma_{X,P}=E[\frac{h_i(X,P)}{f_i(X,P)}]\approx\frac{|P|}{|X|}$. The prompt length $|P|$ is typically much smaller than the instance embedding size, ensuring that $\sigma_{X,P}$ is close to 0. Furthermore, from an optimization perspective, the first term in Eq.~\ref{equ.OrignalFunc_i} could only adjust $f_i(X,P)$ during the backward process. The learning process for downstream datasets may attempt to reduce $f_i(X,P)$ to achieve optimization objectives while partially discarding pre-trained knowledge, which brings a potential risk of overfitting. However, this can be avoided in our method since we keep the coefficient constant in our approach.

By formula, the vanilla instance forwarding process is effectively decoupled into two distinct attention processes:
\begin{equation}
    \mathcal{DA}(X,[X,P])=\mathcal{A}(X,X)+\sigma_{X,P}\mathcal{A}(X,P),
    \label{equ.DAXFunc}
\end{equation}
as shown in the Instance Forwarding part of Fig.~\ref{fig:main_arch}(b).

\noindent \textbf{Prompt Forwarding.} 
It is noteworthy that, despite being non-intuitive in the formula, the original entire self-attention has been decoupled into four \textit{independent} attention subprocesses after decoupling instance forwarding. This necessitates the redefinition of $f$ and $h$ in Eq.~\ref{equ.PFunc}. As in Eq.~\ref{equ.DAXFunc}, we approximate the value by the expectations of each coefficient: $E[f_i(P,X)]\approx\frac{|P|}{|P|+|X|}$ and $E[h_i(P,X)]\approx\frac{|X|}{|P|+|X|}$, denoting them as $\beta_{P,X}$ and $1-\beta_{P,X}$, respectively:
\begin{equation}
    \mathcal{DA}_{Re}(P,[X,P])=\beta_{P,X}\mathcal{A}(P,P)+(1-\beta_{P,X})\mathcal{A}(P,X).
    \label{equ.DARePFunc}
\end{equation}

In our practical experiments, we discovered that the first term in Eq.~\ref{equ.DARePFunc} has a detrimental effect on the final generalization performance. Consequently, we remove this term empirically, resulting in the modified equation:
\begin{equation}
        \mathcal{DA}(P,[X,P]) = \mathcal{A}(P,X),
        \label{equ.DAPfunc}
\end{equation}
as illustrated in the Prompt Forwarding part of Fig.~\ref{fig:main_arch}(b). Upon revisiting the Eq.~\ref{equ.DARePFunc}, we recognize that the second term supplements the information about the current input instance, providing subsequent layer prompts with prior knowledge. While the first term appears less interpretable, we surmise that it may introduce extraneous and uninterpretable information for the subsequent layers.

\noindent \textbf{Language-Conditioned Textual Prompting.} 
Recent studies~\cite{LASP,ProGrad} suggest that maintaining the similarity between learned and handcrafted prompts can lead to more generalizable prompts. Both studies propose auxiliary tasks to achieve this goal, which may introduce additional computational overhead and hyperparameters. Alternatively, a more intuitive approach is to condition textual prompt learning on the handcrafted sentences by combining learnable prompts with template prompts in the input of the text side. However, vanilla prompt learning significantly disrupts the generalizable knowledge carried by the input templates. In this paper, we introduce the language-condition textual prompting, which builds upon our decoupled prompt learning, where input knowledge can be better preserved. Formally, the input to the text encoder can be expressed as
\begin{equation}
\boldsymbol{t}=[\mathrm{P}]_1[\mathrm{~P}]_2 \ldots[\mathrm{P}]_M[\mathrm{Manual\ Prompt}],
\end{equation}
where $[\mathrm{Manual\ Prompt}]$ represents the pre-trained CLIP word embeddings of the handcrafted prompts (e.g., ``\texttt{a photo of a [CLS]}'').

\begin{table*}[t]
  \centering
\small
  \begin{tabular}{l|ccc|ccc|ccc|ccc}
    \toprule
    \multirow{2}{*}{Methods}& \multicolumn{3}{c}{\emph{Average}} & \multicolumn{3}{|c}{ImageNet}
    & \multicolumn{3}{|c}{Caltech101} & \multicolumn{3}{|c}{OxfordPets} \\
    \cmidrule(lr){2-4} \cmidrule(lr){5-7} \cmidrule(lr){8-10} \cmidrule(lr){11-13}
    & Base & New & H & Base & New & H & Base & New & H & Base & New & H \\
    \midrule
    CLIP & 69.34 & 74.22 & 71.70 & 72.43 & 68.14 & 70.22 & 96.84 & 94.00 & 95.40 & 91.17 & 97.26 & 94.12 \\
    CoOp & 82.69 & 63.22 & 71.66 & 76.47 & 67.88 & 71.92 & 98.00 & 89.81 & 93.73 & 93.67 & 95.29 & 94.47 \\
    CoCoOp & 80.47 & 71.69 & 75.83 & 75.98 & 70.43 & 73.10 & 97.96 & 93.81 & 95.84 & 95.20 & 97.69 & 96.43 \\
    ProGrad  & 81.89 & 71.85 & 76.54 & 76.35 & 69.26 & 72.63 & 97.91 & 94.40 & 96.12 & 94.86 & 97.52 & 96.17 \\ 
    ProDA & 81.56 & 72.30 & 76.65 & 75.40 & 70.23 & 72.72 & \textbf{98.27} & 93.23 & 95.68 & \textbf{95.43} & 97.83 & \textbf{96.62} \\
    MaPLe  & 82.28 & 75.14 & 78.55 & 76.66 & 70.54 & 73.47 & 97.74 & 94.36 & 96.02 & \textbf{95.43} & 97.76 & 96.58 \\
    \midrule
    DPL & \textbf{83.42} & \textbf{75.76} & \textbf{79.40} & \textbf{77.17} & \textbf{70.92} & \textbf{73.91} & 98.08 & \textbf{94.54} & \textbf{96.28} & 95.34 & \textbf{97.89} & 96.60 \\
    \bottomrule
  \end{tabular}
  \vspace{.5mm}
  
    \begin{tabular}{l|ccc|ccc|ccc|ccc}
    \toprule
    \multirow{2}{*}{Methods} & \multicolumn{3}{c}{StanfordCars} & \multicolumn{3}{|c}{Flowers102} & \multicolumn{3}{|c}{Food101} & \multicolumn{3}{|c}{FGVCAircraft}\\
    \cmidrule(lr){2-4} \cmidrule(lr){5-7} \cmidrule(lr){8-10} \cmidrule(lr){11-13}
    & Base & New & H & Base & New & H & Base & New & H & Base & New & H \\
    \midrule
    CLIP & 63.37 & \textbf{74.89} & 68.65 & 72.08 & \textbf{77.80} & 74.83 & 90.10 & 91.22 & 90.66 & 27.19 & 36.29 & 31.09 \\
    CoOp & \textbf{78.12} & 60.40 & 68.13 & 97.60 & 59.67 & 74.06 & 88.33 & 82.26 & 85.19 & \textbf{40.44} & 22.30 & 28.75 \\
    CoCoOp & 70.49 & 73.59 & 72.01 & 94.87 & 71.75 & 81.71 & 90.70 & 91.29 & 90.99 & 33.41 & 23.71 & 27.74 \\
    ProGrad & 75.17 & 74.37 & 74.77 & 95.44 & 74.04 & 83.39 & \textbf{90.73} & 91.27 & 91.00 & 38.88 & 31.63 & 34.88\\
    ProDA & 74.70 & 71.20 & 72.91 & \textbf{97.70} & 68.68 & 80.66 & 90.30 & 88.57 & 89.43 & 36.90 & 34.13 & 35.46\\
    MaPLe & 72.94 & 74.00 & 73.47 & 95.92 & 72.46 & 82.56 & 90.71 & \textbf{92.05} & \textbf{91.38} & 37.44 & 35.61 & 36.50 \\
    \midrule
    DPL & 76.16 & 74.86 & \textbf{75.50} & 96.39 & 75.77 & \textbf{84.85} & 90.46 & 91.45 & 90.95 & 40.40 & \textbf{36.43} & \textbf{38.31} \\
    \bottomrule
  \end{tabular}
  \vspace{.5mm}

   \begin{tabular}{l|ccc|ccc|ccc|ccc}
    \toprule
    \multirow{2}{*}{Methods} & \multicolumn{3}{c}{SUN397} & \multicolumn{3}{|c}{DTD} & \multicolumn{3}{|c}{EuroSAT} & \multicolumn{3}{|c}{UCF101} \\
    \cmidrule(lr){2-4} \cmidrule(lr){5-7} \cmidrule(lr){8-10} \cmidrule(lr){11-13} 
    & Base & New & H & Base & New & H & Base & New & H & Base & New & H \\
    \midrule
    CLIP  & 69.36 & 75.35 & 72.23 & 53.24 & 59.90 & 56.37 & 56.48 & 64.05 & 60.03 & 70.53 & 77.50 & 73.85  \\
    CoOp  & 80.60 & 65.89 & 72.51 & 79.44 & 41.18 & 54.24 & 92.19 & 54.74 & 68.69 & 84.69 & 56.05 & 67.46\\
    CoCoOp & 79.74 & 76.86 & 78.27 & 77.01 & 56.00 & 64.85 & 87.49 & 60.04 & 71.21 & 82.33 & 73.45 & 77.64\\
    ProGrad & 80.85 & 74.93 & 77.78 & 77.16 & 54.63 & 63.97 & 88.91 & 53.75 & 67.00 & 84.49 & 74.52 & 79.19\\ 
    ProDA & 78.67 & 76.93 & 77.79 & 80.67 & 56.48 & 66.44 & 83.90 & 66.00 & 73.88 & 85.23 & 71.97 & 78.04\\
    MaPLe & 80.82 & 78.70 & 79.75 & 80.36 & 59.18 & 68.16 & 94.07 & \textbf{73.23} & \textbf{82.35} & 83.00 & 78.66 & 80.77\\ 
    \midrule
    DPL & \textbf{81.11} & \textbf{78.84} & \textbf{79.96} & \textbf{81.48} & \textbf{63.53} & \textbf{71.39} & \textbf{95.62} & 69.31 & 80.37 & \textbf{85.38} & \textbf{79.79} & \textbf{82.49} \\
   \bottomrule
  \end{tabular}
\caption{\textbf{Comparison with the state-of-the-art methods on the base-to-new generalization setting}. This result provides persuasive evidence that DPL possesses a remarkable generalization ability while maintaining the superior adaptation ability of prompt learning methods. `H' refers to the harmonic mean of base and new accuracy, highlighting the trade-off between generalization and adaption.}
\label{tab:cmp_b2n}
\end{table*}

\section{Experiments}
\subsection{Evaluation settings and Datasets}
To verify the generalization ability of our method, we evaluate our model in three settings that are widely utilized in prior works~\cite{CoOp, CoCoOp, ProGrad, ProDA, MaPLe}: base-to-new generalization, cross-dataset transfer, and domain generalization.

\noindent\textbf{Base-to-New Generalization:} For each dataset, we divide the classes equally into two subsets: one serves as the base classes, and the other as the new classes. We train the model on the base classes and evaluate its zero-shot generalization ability on the new classes.

\noindent\textbf{Cross-Dataset Transfer and Domain Generalization:} In both settings, we train the model on the \textit{source} dataset using a few-shot manner and directly test its zero-shot transfer ability and robustness to visual domain shifts on the \textit{target} datasets.

\noindent\textbf{Datasets:} As suggested in prior works~\cite{CoOp, CoCoOp, MaPLe}, we employ $11$ publicly available image classification datasets: ImageNet~\cite{ImageNet} and Caltech101~\cite{Caltech101} for generic object recognition, OxfordPets~\cite{OxfordPets}, StanfordCars~\cite{StanfordCars}, OxfordFlowers~\cite{OxfordFlowers}, Food101~\cite{Food101} and FGVCAircraft~\cite{FGVCAircraft} for fine-grained classification, EuroSAT~\cite{EuroSAT} for satellite recognition, UCF101~\cite{UCF101} for action recognition, DTD~\cite{DTD} for texture classification, and finally SUN397~\cite{SUN397} for scene recognition. All $11$ datasets are used in the base-to-new setting. We utilize ImageNet as the \textit{source} dataset and the other $10$ datasets as \textit{target} datasets for cross-dataset transfer. For domain generalization, we use ImageNet as \textit{source} dataset and select ImageNetV2~\cite{ImageNetV2}, ImageNet-Sketch~\cite{ImageNet-Sketch}, ImageNet-A~\cite{ImageNet-A} and ImageNet-R~\cite{ImageNet-R} as \textit{target} datasets.

\begin{table*}[t]
    \small
    \begin{tabular}{l c | ccccccccccc}
    \toprule
    & \rotbox{ImageNet} & \rotbox{Caltech101} & \rotbox{Pets} & \rotbox{Cars} & \rotbox{Flowers} & \rotbox{Food101} & \rotbox{Aircraft} & \rotbox{SUN397} & \rotbox{DTD} & \rotbox{EuroSAT} & \rotbox{UCF101} & \rotbox{\emph{Avg}} \\
    \midrule
    CLIP &  66.73 & 93.35 & 88.25 & 65.48 & 67.44 & 83.65 & 23.67 & 62.59 & 44.27 & 42.01 & 65.13 & 63.58 \\
    CoOp & \textbf{71.51} & 93.70 & 89.14 & 64.51 & 68.71 & 85.30 & 18.47 & 64.15 & 41.92 & 46.39 & 66.55 & 63.88 \\
    CoCoOp & 71.02 & \textbf{94.43} & 90.14 & 65.32 & 71.88 & 86.06 & 22.94 & 67.36 & 45.73 & 45.37 & 68.21 & 65.74 \\
    MaPLe & 70.72 & 93.53 & 90.49 & 65.57 & \textbf{72.23} & 86.20 & \textbf{24.74} & 67.01 & \textbf{46.49} & 48.06 & 68.69 & 66.30 \\
    \midrule
    DPL & 70.77 & 93.74 & \textbf{90.67} & \textbf{65.64} & 71.67 & \textbf{86.36} & 24.67 & \textbf{67.46} & 45.76 & \textbf{52.69} & \textbf{70.10} & \textbf{66.88} \\
    \bottomrule
    \end{tabular}
    \caption{\textbf{Comparison of prompt learning methods in cross-dataset transfer setting}. Our method achieves comparable results with state-of-the-art methods on 10 datasets and attains the highest average accuracy, demonstrating its remarkable performance in zero-shot transfer.}
     \label{tab:cmp_cross}
\end{table*}

\subsection{Implementation Details}
Our experiments are conducted on CLIP, based on ViT-B/16, and all results are averaged over three seeds. We employ stochastic gradient descent as the optimizer, following prior works~\cite{CoOp, CoCoOp, MaPLe}. All experiments are performed on a single A100 GPU.

Unless otherwise specified, our method is based on \textit{multi-modal prompt learning}, where learnable prompts are incorporated into both the visual and textual sides of CLIP. In the base-to-new setting, we set the visual and textual prompt depth to $9$ and set the prompt length per layer to $8$ and $4$, respectively. All models are trained for 10 epochs with a batch-size of 4. The learning rate is set to $0.1$ for visual prompts following~\cite{ProVP}, and $0.002$ for textual prompts following~\cite{CoOp, CoCoOp}. For cross-dataset and domain generalization, where ImageNet is used as the training set, we maintain the same setting for the textual side and reduce the visual depth and prompt lengths per layer to $3$ and $2$, respectively, as both benchmarks rely less on visual priors. The learning rate is set to $0.003$ for both visual and textual prompts. We initialize the visual prompts using Xavier~\cite{Xavier} following~\cite{VPT}, and randomly initialize the textual prompts from a normal distribution following~\cite{CoOp, MaPLe}, except for the first layer, which we initialize from the pre-trained CLIP word embeddings of ``\texttt{a photo of a}'' following~\cite{MaPLe}. For further experimental details and hyperparameter settings, please refer to the supplementary material.

\subsection{Main Results}
\noindent\textbf{Base-to-New Generalization.}
In this experimental setup, we present the accuracy of the base and new categories separately, as well as their harmonic mean, which quantifies the trade-off between the two metrics. In Tab.~\ref{tab:cmp_b2n}, we compare our DPL with previous approaches, including zero-shot CLIP~\cite{CLIP}, CoOp~\cite{CoOp}, CoCoOp~\cite{CoCoOp}, ProGrad~\cite{ProGrad}, ProDA~\cite{ProDA}, and MaPLe~\cite{MaPLe}. DPL achieves state-of-the-art performance on both base and new categories, resulting in the highest harmonic mean of $79.40\%$. Notably, our method only uses $72$K trainable parameters, which is $48\times$ fewer than MaPLe's $3.39$M parameters.

Previous prompt learning methods often suffer from overfitting to seen categories, resulting in poor performance on unseen categories. Prior studies attempt to alleviate such performance drops by introducing extra network modules~\cite{MaPLe,CoCoOp} or auxiliary regularization tasks~\cite{ProGrad}. Both approaches may increase the computation overhead and make the optimization process complicated. In contrast, our method demonstrates superior transfer and generalization performance without introducing any additional trainable parameters or hyperparameters. Specifically, DPL achieves the best performance on 7 out of 11 datasets on new classes and outperforms all previous prompt learning methods on 9 out of 11 datasets, providing strong evidence for the superior generalizability of our proposed decoupled prompt learning.

\noindent\textbf{Cross-Dataset Transfer.} Tab.~\ref{tab:cmp_cross} presents the results of cross-dataset transfer. DPL achieves comparable performance with previous state-of-the-art methods. Specifically, DPL outperforms all previous methods on 6 out of 10 datasets. Furthermore, our method achieves the best average performance across all datasets, demonstrating the improvement of the zero-shot transfer ability.

\begin{table}[t]
    \small
    \begin{tabular}{l c | cccc}
    \toprule
    & ImageNet & -V2 & -S & -A & -R \\
    \midrule
    CLIP & 66.73 & 60.83 & 46.15 & 47.77 & 73.96 \\
    CoOp & \textbf{71.51} & \textbf{64.20} &  47.99 & 49.71 & 75.21\\
    CoCoOp & 71.02 & 64.07 & 48.75 & 50.63 & 76.18 \\
    MaPLe & 70.72 & 64.07 & 49.15 & 50.90 & 76.98 \\
    \midrule
    DPL & 70.77 & 64.10 & \textbf{49.50} & \textbf{50.92} & \textbf{77.23}\\
    \bottomrule
    \end{tabular}
    \caption{\textbf{Comparison in domain generalization setting.} Our method consistently achieves the highest performance, demonstrating its superior robustness to out-of-distribution scenarios.}
    \label{tab:domain_gen}
\end{table}

\noindent\textbf{Domain Generalization.}
The results of domain generalization are presented in Tab.~\ref{tab:domain_gen}. Our method consistently achieves the best performance on out-of-distribution datasets. These findings demonstrate that our method effectively improves the robustness and generalization of the learned prompts.

\subsection{Ablation Studies}

\noindent\textbf{Stepwise Ablation of DPL.} Tab.~\ref{tab:stepwise_ablation} presents the results of the stepwise ablation analysis of our proposed method. We choose the \textit{vanilla multi-modal prompt learning} (short as MPL) as our baseline, which has the same hyperparameters and training settings as our approach. We gradually decouple the self-attention process within $[X,P]$ (i.e., combining Eq.~\ref{equ.DAXFunc} and Eq.~\ref{equ.DARePFunc}), remove the self-attention within prompt tokens $P$ (i.e., combining Eq.~\ref{equ.DAXFunc} and Eq.~\ref{equ.DAPfunc}), and condition the textual prompt learning on specified template prompts.

We observe a significant improvement of $2.09\%$ in new accuracy achieved by decoupling the attention of the instance and prompt tokens, which highlights the original computational structure of zero-shot CLIP, and ensure that the rich generalizable pre-trained knowledge remains uncompromised when transferring to downstream tasks. Furthermore, removing the self-attention within prompts leads to a further improvement of $0.42\%$, demonstrating the redundancy of the interactions between prompts. Finally, language-conditioned textual prompting naturally preserves the generalizable ability of the text-side input, resulting in a $0.83\%$ improvement.

Additionally, it is worth noticing that other approaches with similar generalization performance, such as MaPLe, provide much less improvement than we achieve with language-conditioned prompts. This demonstrates the compatibility of LCTP with DPL, where the input knowledge in handcrafted templates is much less disrupted.

\begin{table}[h]
  \centering
  \small
  \begin{tabular}{l|ccc|ccc}
    \toprule
    Models & DA & SR & LC &Base & New & H \\
    \midrule
    MaPLe &  &  & & 82.28 & 75.14 & 78.55 \\
    MaPLe &  &  & \checkmark  & 82.26 & 75.30~(\textit{+0.16}) & 78.63\\
    \midrule
    MPL &  & & & 83.48 & 72.42 & 77.55\\
     & \checkmark & & & 83.21 & 74.51 (+\textit{2.09}) & 78.62\\
     & \checkmark & \checkmark & & 83.15 & 74.93 (+\textit{0.42}) & 78.83\\
    DPL & \checkmark & \checkmark & \checkmark & 83.42 & 75.76 (+\textit{0.83}) & 79.40\\
    \bottomrule
  \end{tabular}
  \caption{\textbf{Stepwise ablation analysis of our proposed methods}, including Decoupled Attention (DA), Prompt Self-attention Removing (SR), and Language-Conditioned Textual Prompting (LC), demonstrates that our methods consistently and significantly improve the generalization performance while maintaining transferability. Furthermore, we find that LC, when combined with our proposed DA, can better preserve the generalization knowledge, achieving an improvement of $0.83$ compared to MaPLe's $0.16$.} 
  \label{tab:stepwise_ablation}
\end{table}

\begin{table}[h]
  \centering
  \small
  \begin{tabular}{lcccc}
    \toprule
    Methods & \# Params (K) & Base & New & H \\
    \midrule
    CoOp & 2 & 82.69 & 63.22 & 71.66\\
    CoCoOp & 34.52 & 80.47 & 71.69 & 75.83\\
    ProDA & 256 & 81.56 & 72.30 & 76.65\\
    MaPLe & 3471.75 & 82.28 & 75.14 & 78.55\\
    \midrule
    DPL & 72 & \textbf{83.42} & 75.76 & \textbf{79.40}\\
    DPL$_{\text{layers}/2}$ & 32 & 81.11 & 75.19 & 78.04\\
    DPL$_{\text{tokens}/2}$ & 36 & 80.81 & \textbf{76.28} & 78.48\\
    \bottomrule
  \end{tabular}
\caption{\textbf{Ablation of trainable parameters.} We experiment with two manners to reduce the number of parameters: halving the number of layers where the prompt is inserted and halving the number of prompts inserted at each layer. Despite the reduction in parameters, our approach maintains excellent generalization performance.}
\label{tab:ablation_param}
\end{table}

\noindent\textbf{Ablation of Trainable Parameters.} The number of trainable parameters has always been an important metric for the efficiency of prompt learning methods. Tab.~\ref{tab:ablation_param} compares the number of trainable parameters in DPL with existing prompt learning approaches, while also evaluating the effect of reducing the number of parameters on our method. Our approach is the most parameter-efficient one with $3.5\times$ fewer parameters than ProDA and $48\times$ fewer parameters than MaPLe. Moreover, to further understand the relationship between parameters and performance in DPL, we experimented with two manners of halving the number of parameters: (1) halving the number of layers in which prompt tokens are inserted, i.e., from $1\rightarrow 9$ to $1 \rightarrow 4$; (2) halving the number of prompt tokens in each layer. Despite the reduction in the number of parameters, our method still exhibits strong generalization to unseen categories, with an average accuracy that surpasses all existing approaches.

\begin{figure}[t]
    \centering
    \includegraphics[width=0.42\textwidth]{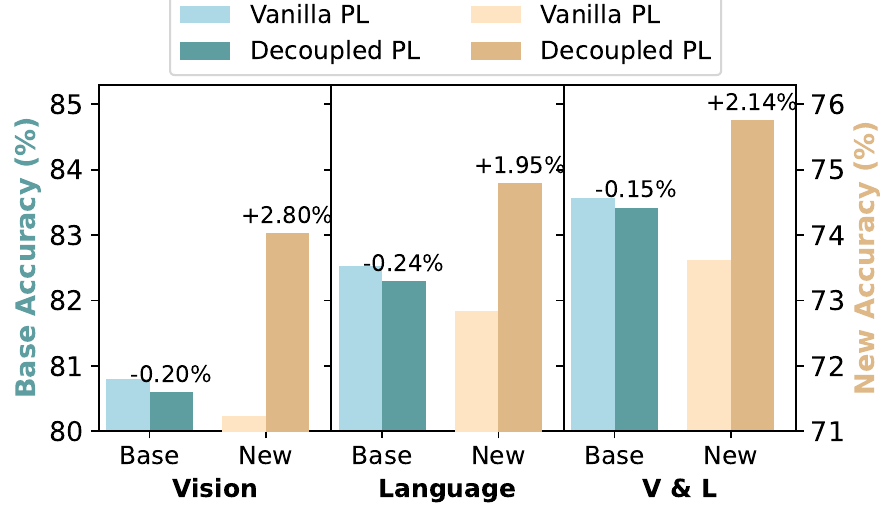}
    \caption{\textbf{Ablation of DPL in different modalities.} To better demonstrate the effects of decoupled prompt learning, we uniformly employ LCTP. Our approach significantly enhances generalization when applied to all three settings, demonstrating the universal applicability of our proposed method.}
    \label{fig:ablation_modality}
\end{figure}

\noindent\textbf{DPL is Universally Effective Across Modalities.} We verify the effectiveness of DPL in visual, textual, and visual-textual multimodal settings. To better demonstrate the effects of decoupling attention, we uniformly use language-conditioned textual prompt learning in all experiments except for the single visual modal setting. As shown in Fig.~\ref{fig:ablation_modality}, DPL consistently and significantly improves the generalization ability of prompt learning across different modalities by 2.8\% in visual, 1.95\% in textual, and 2.14\% in visual-textual modality. These results highlight the generality of our approach across different modalities.

\begin{figure}[t]
    \centering
    \includegraphics[width=0.35\textwidth]{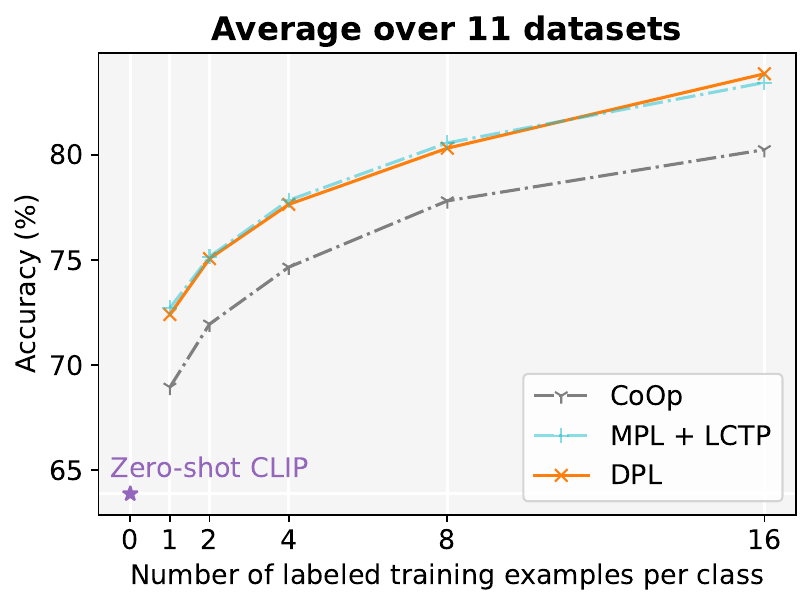}
    \caption{ \textbf{Few-shot average accuracy over 11 datasets.} This provides convincing evidence that DPL significantly enhances the generalizability of vanilla prompt learning while maintaining its superior adaptation capability.}
    \label{fig:ablation_few_shot}
\end{figure}

\noindent\textbf{Whether DPL Sacrifices Transfer Capability.} 
In Tab.~\ref{tab:stepwise_ablation} and Fig.~\ref{fig:ablation_modality}, we observe a minor decrease in the base accuracy of our method. This motivates us to investigate whether DPL sacrifices transfer performance compared to vanilla prompt learning. Few-shot learning is an important and representative benchmark to test the model's adaptation ability. We follow previous works \cite{CoOp} to train the model
on 1, 2, 4, 8, and 16 shots, and evaluate the trained model on the full test sets. The average accuracy across the 11 datasets is shown in Fig.~\ref{fig:ablation_few_shot}. We observe that DPL is slightly lower than the baseline (vanilla multi-modal prompt learning with language-conditioned textual prompting, short as MPL+LCTP) for 1, 4, and 8 shots, while marginally outperforming it for 16 shots. Overall, DPL still maintains the excellent transfer ability of prompt learning methods.


\section{Conclusion}
In this paper, we have unveiled the fact that the original attention operation in prompt learning involves four distinct sub-processes. We analyze the potential effectiveness of the separate sub-processes and reveal that each of them is responsible for distinct objectives, and certain terms can be reinforced to bolster the robustness and generalization ability. Based on these insights, we proposed a novel approach called Decoupled Prompt Learning by decoupling the original attention, and achieves a more robust and effective optimization process for prompts by the well-designed recombination of separated sub-processes. Remarkably, our method has achieved the state-of-the-art results on three representative benchmarks without any auxiliary regularization task or additional module. Our improvement in prompt learning provides a strong and promising baseline for adapting VLMs. We hope it will facilitate research in this field on the generalizability of adapting foundational models.

\bibliography{aaai24}
\newpage


\section{Supplementary Material}

\subsection{Detailed Derivation of the Attention Reformulation}

Here we present a comprehensive formula derivation for the separate results of the attention process within the vanilla prompt learning. Utilizing the definitions provided in the main paper and accounting for the constant term as $\sqrt{D}$, we obtain the following equation:
\begin{equation}
    \small
    \begin{aligned}
        \mathcal{A}([X,P],[X,P])=[\mathcal{A}(X,[X,P]),\mathcal{A}(P,[X,P])],
    \end{aligned}
\end{equation}
as the original instance and prompt forwarding processes respectively. And for a token embedding $x_i\in X$, we have:
\begin{equation}
    \small
    \begin{aligned}
        \mathcal{A}_i(X,[X,P])&=\frac{\sum\limits_{y_j\in [X,P]}e^{q(x_i)k(y_j)}v(y_j)}{\lambda_i(X,[X,P])}\\
        &=\frac{\sum\limits_{x_j\in X}e^{q(x_i)k(x_j)}v(x_j)+\sum\limits_{p_j\in P}e^{q(x_i)k(p_j)}v(p_j)}{\lambda_i(X,[X,P])}\\
        &=\frac{\lambda_i(X,X)\mathcal{A}_i(X,X)+\lambda_i(X,P)\mathcal{A}_i(X,P)}{\lambda_i(X,[X,P])}\\
        &=f_i(X,P)\mathcal{A}_i(X,X)+h_i(X,P)\mathcal{A}_i(X,P).
    \end{aligned}
    \label{eq.deriv}
\end{equation}
Noticing that $\mathcal{A}(P,[X,P])=\mathcal{A}(P,[P,X])$. By following a similar derivation process as Eq.~\ref{eq.deriv}, we can deduce the result of prompt forwarding as presented in Eq.~6.
\subsection{Additional Implementation Details}
Following the previous researches~\cite{CoOp, CoCoOp, ProGrad, MaPLe}, we implement random resized cropping and flipping into the training process. Additionally, we employ the warm-up technique, whereby the learning rate is initially fixed at $1e-5$ for the first epoch before gradually decaying according to the cosine annealing rule from the initial value.  In the context of Language-Conditioned Textual Prompting, we have designed specified templates for the text-side input, the majority of which are sourced from CLIP~\cite{CLIP}. These templates are itemized in Tab.~\ref{tab:template} for your reference.
\begin{table}[h]
\centering
\footnotesize
\begin{tabular}{l|l}
\toprule
\textbf{Caltech101} & \textit{a photo of a [CLS], a type of rendition.} \\ \cmidrule(lr){1-2}

\textbf{OxfordPets} & \textit{a photo of a [CLS], a type of pet.} \\ \cmidrule(lr){1-2}

\textbf{StanfordCars} &  \textit{a photo of a [CLS].} \\ \cmidrule(lr){1-2} 

\textbf{Flowers102} & \textit{a photo of a [CLS], a type of flower.}\\  \cmidrule(lr){1-2}

\textbf{Food101} & \textit{a type of food, a photo of [CLS].}\\ \cmidrule(lr){1-2}

\multirow{2}{*}{\textbf{FGVCAircraft}} & \textit{a photo of a [CLS],} \\ 

& \textit{ which is a type of an aircraft.}\\ \cmidrule(lr){1-2}

\textbf{SUN397} & \textit{a photo of a [CLS].} \\ \cmidrule(lr){1-2}

\textbf{DTD} & \textit{a photo of a [CLS], a type of a texture.} \\ \cmidrule(lr){1-2}

\textbf{EuroSAT} &  \textit{a centered satellite photo of [CLS].} \\ \cmidrule(lr){1-2}

\textbf{UCF101} &  \textit{a photo of a person doing [CLS].}\\  \cmidrule(lr){1-2}

\textbf{ImageNet} & \textit{a photo of a [CLS].} \\ \cmidrule(lr){1-2}

\textbf{ImageNetV2} &\textit{a photo of a [CLS].} \\ \cmidrule(lr){1-2}

\textbf{ImageNet-Sketch} & \textit{a photo of a [CLS], a type of sketch. }\\ \cmidrule(lr){1-2}
 
\multirow{2}{*}{\textbf{ImageNet-A}} &  \textit{a photo of a [CLS],} \\ 
 & \textit{ which is natural adversarial.}\\  \cmidrule(lr){1-2}

\textbf{ImageNet-R} &  \textit{a photo of a [CLS], a type of rendition.} \\  \bottomrule
\end{tabular}
\vspace{-2mm}
\caption{\textbf{Handcrafted templates used in Language-Conditioned Textual Prompting.}}
\label{tab:template}
\vspace{-3mm}
\end{table}

\subsection{DPL in Extreme Few-shot Scenarios}
\begin{figure}[htbp]
    \centering
    \includegraphics[width=0.42\textwidth]{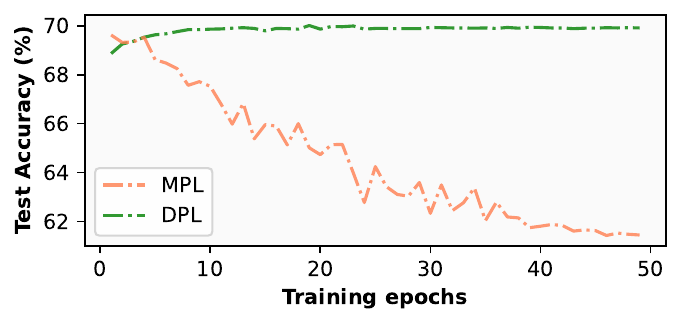}
    \vspace{-3mm}
    \caption{\textbf{Record of the training epochs and corresponding testing accuracy on ImageNet in a 1-shot setting.} }
    \label{fig:1shot-overfit}
    \vspace{-4mm}
\end{figure}

During our experiment, we made an intriguing observation (e.g., Fig.~\ref{fig:1shot-overfit}): 
when confronted with extremely few-shot learning scenarios, vanilla prompt learning methods are prone to overfitting the training set, thereby impairing its performance on the test set. While this issue can be mitigated by employing an additional validation set, it is often challenging to collect sufficient data under this paradigm. 

However, we note that our proposed Decoupled Prompt Learning (DPL) effectively curbs the overfitting phenomenon. We hypothesize that this is partially due to our decoupled attention mechanism, which avoids redundant computations and unnecessary interactions. Such reductions can prevent the model from focusing too much on specific input 
instances from the training set, and instead allow it to prioritize task-specific information. Further investigation into this intriguing phenomenon will be deferred to future research.

\subsection{Attention Map Visualization}

\begin{figure}[h]
    \centering
    \includegraphics[width=0.42\textwidth]{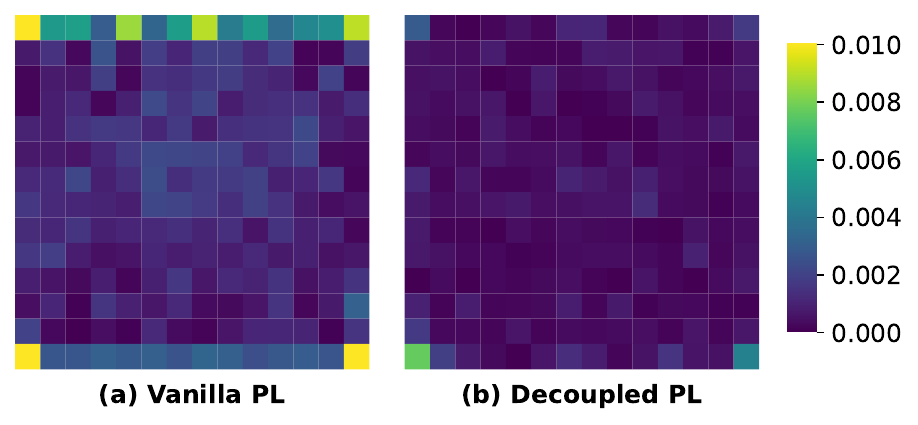}
    \vspace{-3mm}
    \caption{
    \textbf{Visualization of the distance between the attention map generated by two prompt learning methods (vanilla and DPL) and zero-shot CLIP}. Each value of the attention map is calculated using the `\texttt{[cls]}' token as the query, instance embeddings as the key, and averaging over the entire ImageNet test set.}
    \label{fig:attnmap}
\end{figure}

As we recombine the computation processes of $\mathcal{A}(X,X)$, we empirically analyze the impact of the different prompt learning methods on the attention map within instance embeddings. Specifically, we focused on the last transformer layer, which is closest to the final output. To calculate each value of the attention map, we utilized the `[\texttt{CLS}]' token as the query, and all instance embeddings as the key. We then visualize the differences in the attention map between the prompt learning method and the zero-shot CLIP from the visual side, as depicted in Fig.~\ref{fig:attnmap}. Our results demonstrate that DPL preserves the correlations between input instance embeddings in the last layer much better than vanilla prompt learning. This empirical finding indicates that the original computation structure of the zero-shot CLIP model is well-preserved in earlier layers by DPL, which contributes to our superior generalization performance.

\subsection{DPL on Different Backbone Scales}

\begin{table}[h]
  \centering
  \small
  \vspace{-2mm}
  \begin{tabular}{lcccc}
    \toprule
    Methods & Backbone & Base & New & H \\
    \midrule
    CLIP & \multirow{3}{*}{ViT-B/32} & 67.23& 71.80 & 69.44 \\
    MPL$^*$ &  & 79.34 & 67.60 & 73.00 \\
    DPL &  & 79.78 & 71.09(+\textit{3.49}) & 75.18 \\
    \midrule
    CLIP & \multirow{3}{*}{ViT-B/16} & 69.34 & 74.22 & 71.70 \\
    MPL$^*$ &  & 83.57 & 73.62 & 78.28 \\
    DPL &  & 83.42 & 75.76(+\textit{2.14}) & 79.40\\
    \midrule
    CLIP & \multirow{3}{*}{ViT-L/14} &76.65 & 80.34 & 78.45 \\
    MPL$^*$ &  & 86.47 & 77.10 & 81.52 \\
    DPL &  & 86.72 & 81.21(+\textit{4.11}) & 83.88 \\
    \bottomrule
  \end{tabular}
  \vspace{-2mm}
\caption{\textbf{Ablation of our proposed DPL on various backbone scales.} We compare DPL with vanilla multi-modal prompt learning (MPL) equipped with LCTP, denoted as MPL$^*$.} 
\label{tab:ablation_backbone}
\vspace{-4mm}
\end{table}

In addition to the ablation studies across different modalities, we have further validated the versatility of our proposed DPL by conducting ablation experiments on two other backbones scales: ViT-B/32 based CLIP and ViT-L/14 based CLIP~\cite{CLIP}. In the Base/32 case, we reduce the number of visual prompts per layer to 4 to accommodate the halving of the number of input visual embeddings. For the Large/14 case, we increase the number of visual prompts per layer to 9 and the number of layers for prompt insertion to 18. Additionally, we lower the learning rate of the visual side to $0.02$ to stabilize the training process.

As illustrated in Tab.~\ref{tab:ablation_backbone}, our method yields significant improvements in the accuracy of new categories while maintaining the base performance. Specifically, DPL achieves a $3.49\%$ and $4.11\%$ increase in accuracy on the Base/32 and Large/14 scales, respectively, leading to a substantial improvement in the harmonic mean. These results provide further compelling evidence of the versatility and effectiveness of our approach. Remarkably, as DPL shows a greater improvement of the performance on new classes when the size of the foundational model grows, it indicates that our method holds the potential to be extended to more powerful models, and could enhance the generalization ability in even more challenging tasks. 

\begin{table*}[h]
  \centering
\small

  \begin{tabular}{l|ccc|ccc|ccc|ccc}
    \toprule
    \multirow{2}{*}{Methods}& \multicolumn{3}{c}{\emph{Average}} & \multicolumn{3}{|c}{ImageNet}
    & \multicolumn{3}{|c}{Caltech101} & \multicolumn{3}{|c}{OxfordPets} \\
    \cmidrule(lr){2-4} \cmidrule(lr){5-7} \cmidrule(lr){8-10} \cmidrule(lr){11-13}
    & Base & New & H & Base & New & H & Base & New & H & Base & New & H \\
    \midrule
    MaPLe  & 82.28 & 75.14 & 78.55 & 76.66 & 70.54 & 73.47 & 97.74 & 94.36 & 96.02 & 95.43 & 97.76 & 96.58 \\
    MaPLe$^*$  & 82.26 & 75.30 & 78.63 & 76.66 & 70.51 & 73.46 & 97.98 & 93.71 & 95.80 & 95.62 & 97.86 & 96.73 \\

    \midrule
    MPL & 83.48 & 72.42 & 77.55 & 77.29 & 68.56 & 72.66 & 98.21 & 93.19 & 95.63 & 95.21 & 97.33 & 96.26 \\
    MPL + DA & 83.21 & 74.51 & 78.62 & 77.14 & 70.39 & 73.61 & 98.17 & 94.98 
  & 96.55  & 95.18 & 97.41 & 96.28 \\
    MPL + DASR & 83.15 & 74.93 & 78.83 & 77.24 & 70.51 & 73.72 & 98.15 & 94.94 & 96.52 & 95.52 & 97.58 & 96.54 \\
    \midrule
    DPL$_{\text{layers}/2}$ & 81.11 & 75.19 & 78.04 & 76.29 & 70.67 & 73.37 & 97.87 & 94.40 & 96.10 & 95.22 & 97.37 & 96.28 \\
    DPL$_{\text{tokens}/2}$ & 73.65 & 74.89 & 74.26 & 93.37 & 75.60 & 83.55 & 90.73 & 91.57 & 91.15 & 36.77 & 35.97 & 36.37 \\
    \midrule

    VPL$^*$ & 80.80 & 71.23 & 75.72 & 75.80 & 68.41 & 71.92 & 98.04 & 93.23 & 95.57 & 95.22 & 95.55 & 95.38 \\
    VPL$^*$ + DASR & 80.60 & 74.03 & 77.18 & 75.70 & 69.42 & 72.42 & 98.06 & 93.67 & 95.81 & 95.27 & 96.65 & 95.96 \\
    TPL$^*$ &  82.53 &	72.84 &	77.38 &	76.68 &	68.53 &	72.38 &	98.24 &	94.87 &	96.53 &	94.86 &	97.33 &	96.08  \\
    TPL$^*$ + DASR & 82.09 & 74.79 &	78.27 &	76.69 &	71.04 &	73.76 &	98.11 &	94.83 &	96.44 & 95.32 &	97.62 &	96.46 \\
    MPL$^*$ &  83.57 &	73.62 &	78.28 &	77.05 &	69.10 &	72.86 &	98.54 &	93.92 &	96.17 &	95.71 &	97.48 &	96.59 \\
    MPL$^*$ + DASR & 83.42 &	75.76 &	79.40 &	77.17 &	70.92 &	73.91 &	98.08 &	94.54 &	96.28 &	95.34 &	97.89 &	96.60 \\
    \midrule
    
    DPL & 83.42 & 75.76 & 79.40 & 77.17 & 70.92 & 73.91 & 98.08 & 94.54 & 96.28 & 95.34 & 97.89 & 96.60 \\
    \bottomrule
  \end{tabular}
  \vspace{.6mm}

    \begin{tabular}{l|ccc|ccc|ccc|ccc}
    \toprule
    \multirow{2}{*}{Methods} & \multicolumn{3}{c}{StanfordCars} & \multicolumn{3}{|c}{Flowers102} & \multicolumn{3}{|c}{Food101} & \multicolumn{3}{|c}{FGVCAircraft}\\
    \cmidrule(lr){2-4} \cmidrule(lr){5-7} \cmidrule(lr){8-10} \cmidrule(lr){11-13}
    & Base & New & H & Base & New & H & Base & New & H & Base & New & H \\
    \midrule
    MaPLe & 72.94 & 74.00 & 73.47 & 95.92 & 72.46 & 82.56 & 90.71 & 92.05 & 91.38 & 37.44 & 35.61 & 36.50 \\
    MaPLe$^*$  & 72.41 & 73.90 & 73.15 & 96.11 & 73.19 & 83.10 & 90.51 & 91.97 & 91.23 & 37.05 & 35.81 & 36.42 \\

    \midrule
    MPL & 77.50 & 71.97 & 74.63 & 97.53 & 71.63 & 82.60 & 90.05 & 91.15 & 90.60 & 38.12 & 33.89 & 35.88 \\
    MPL + DA & 76.10 & 74.34 & 75.21 & 96.74 & 73.45 & 83.50 & 90.53 & 91.38 & 90.95 & 39.06 & 34.91 & 36.87 \\
    MPL + DASR & 75.96 & 74.48 & 75.21 & 96.55 & 74.04 & 83.81 & 90.53 & 91.45 & 90.99 & 39.36 & 34.51 & 36.78 \\
    \midrule
    DPL$_{\text{layers}/2}$ & 73.65 & 74.89 & 74.26 & 93.37 & 75.60 & 83.55 & 90.73 & 91.57 & 91.15 & 36.77 & 35.97 & 36.37 \\
    DPL$_{\text{tokens}/2}$ & 72.14 & 75.70 & 73.88 & 91.82 & 76.38 & 83.39 & 90.74 & 91.67 & 91.20 & 35.43 & 37.51 & 36.44 \\
    \midrule

    VPL$^*$ & 71.55 & 71.82 & 71.68 & 91.00 & 68.89 & 78.42 & 90.01 & 90.70 & 90.35 & 34.75 & 33.07 & 33.89 \\
    VPL$^*$ + DASR & 71.01 & 73.20 & 72.09 & 90.73 & 72.72 & 80.73 & 90.41 & 90.71 & 90.56 & 34.55 & 35.51 & 35.02 \\
    TPL$^*$ & 75.10 &	72.05 &	73.54 &	97.41 &	73.78 &	83.96 &	89.53 &	91.09 &	90.30 &	39.28 &	34.23 &	36.58  \\
    TPL$^*$ + DASR & 75.48 &	75.28 &	75.38 &	95.82 &	76.05 &	84.80 &	90.66 &	91.74 &	91.20 &	38.82 &	36.19 &	37.46  \\
    MPL$^*$ &  77.05 &	73.76 &	75.37 &	97.34 &	73.26 &	83.60 &	90.37 &	91.35 &	90.86 &	39.70 &	33.81 &	36.52  \\
    MPL$^*$ + DASR & 76.16 &	74.86 &	75.50 &	96.39 &	75.77 &	84.85 &	90.46 &	91.45 &	90.95 &	40.40 &	36.43 &	38.31 \\
    \midrule
    
    DPL & 76.16 & 74.86 & 75.50 & 96.39 & 75.77 & 84.85 & 90.46 & 91.45 & 90.95 & 40.40 & 36.43 & 38.31 \\
    \bottomrule
  \end{tabular}
  \vspace{.6mm}

   \begin{tabular}{l|ccc|ccc|ccc|ccc}
    \toprule
    \multirow{2}{*}{Methods} & \multicolumn{3}{c}{SUN397} & \multicolumn{3}{|c}{DTD} & \multicolumn{3}{|c}{EuroSAT} & \multicolumn{3}{|c}{UCF101} \\
    \cmidrule(lr){2-4} \cmidrule(lr){5-7} \cmidrule(lr){8-10} \cmidrule(lr){11-13} 
    & Base & New & H & Base & New & H & Base & New & H & Base & New & H \\
    \midrule
    MaPLe & 80.82 & 78.70 & 79.75 & 80.36 & 59.18 & 68.16 & 94.07 & 73.23 & 82.35 & 83.00 & 78.66 & 80.77\\
    MaPLe$^*$  & 80.47 & 78.13 & 79.28 & 80.29 & 60.83 & 69.22 & 94.11 & 73.95 & 82.82 & 83.68 & 78.46 & 80.99 \\

    \midrule
    MPL & 81.67 & 77.05 & 79.29 & 82.25 & 50.89 & 62.88 & 95.25 & 65.94 & 77.93 & 85.18  & 74.98 & 79.76 \\
    MPL + DA & 80.46 & 78.89 & 79.67 & 81.36 & 60.87 & 69.64 & 95.39 & 65.57 & 77.72 & 85.23 & 77.41 & 81.13 \\
    MPL + DASR & 80.65 & 79.11 & 79.87 & 81.60 & 61.75 & 70.30 & 94.33 & 68.07 & 79.08 & 84.76 & 77.77 & 81.11 \\
    \midrule
    DPL$_{\text{layers}/2}$ & 79.43 & 78.68 & 79.05 & 78.58 & 64.41 & 70.79 & 86.31 & 63.79 & 73.36 & 83.97 & 79.74 & 81.80 \\
    DPL$_{\text{tokens}/2}$ & 78.21 & 78.43 & 78.32 & 77.77 & 65.62 & 71.18 & 91.24 & 69.58  & 78.95 & 82.42 & 80.89 & 81.65 \\
    \midrule

    VPL$^*$ & 79.81 & 76.35 & 78.04 & 79.98 & 52.09 & 63.09 & 88.36 & 60.36 & 71.72 & 84.30 & 73.09 & 78.30 \\
    VPL$^*$ + DASR & 78.03 & 78.10 & 78.06 & 73.54 & 58.29 & 65.03 & 95.74 & 69.21 & 80.34 & 83.52 & 76.89 & 80.07 \\
    TPL$^*$ & 81.72 &	76.53 &	79.04 &	80.17 &	55.84 &	65.83 &	90.17 &	61.92 &	73.42 &	84.62 &	75.10 &	79.58  \\
    TPL$^*$ + DASR & 80.82 &	78.66 &	79.73 &	79.44 &	60.51 &	68.69 &	88.86 &	61.56 &	72.73 &	83.01 &	79.16 &	81.04 \\
    MPL$^*$ & 81.68 &	77.45 &	79.51 &	81.29 &	56.68 &	66.79 &	94.30 &	64.68 &	76.73 &	86.20 &	78.28 &	82.05 \\
    MPL$^*$ + DASR & 81.11 &	78.84 &	79.96 &	81.48 &	63.53 &	71.39 &	95.62 &	69.31 &	80.37 &	85.38 &	79.79 &	82.49  \\
    \midrule
    
    DPL & 81.11 & 78.84 & 79.96 & 81.48 & 63.53 & 71.39 & 95.62 & 69.31 & 80.37 & 85.38 & 79.79 & 82.49 \\
   \bottomrule
  \end{tabular}
\caption{\textbf{Detailed ablation results in base-to-new setting across 11 datasets.} `*' denotes the model equipped with LCTP, `DA' represents the Decoupled Attention, and `DASR' refers to the Decoupled Attention with Self-attention within prompts Removed.}
\label{tab:detail_ablation}
\end{table*}
\subsection{Detailed Ablation Results}

Due to space limitations in the main paper, we have only presented the average results across 11 datasets in our ablation study. For the convenience of future studies, we include the detailed results in Tab.~\ref{tab:detail_ablation}.

\bigskip

\end{document}